\pdfoutput=1

\documentclass[11pt]{article}

\usepackage{EACL2023}

\usepackage{times}
\usepackage{latexsym}

\usepackage[T1]{fontenc}

\usepackage[utf8]{inputenc}

\usepackage{microtype}

\usepackage{inconsolata}

\usepackage{hyperref}
\usepackage{url}

\usepackage{graphicx}
\usepackage{amsfonts}
\usepackage{amsmath}
\usepackage{amssymb}
\usepackage{pifont}
\usepackage{bm}
\usepackage{nicefrac}
\usepackage{multirow}
\usepackage{subfigure}
\usepackage[font=small]{caption}
\usepackage{soul}
\usepackage{xcolor}
\usepackage{xspace}
\usepackage{booktabs}
\usepackage{comment}
\usepackage{sidecap}

\newcommand{\eplabel}{{CoLaCTC}\xspace}
\newcommand{\etru}{Tru\xspace}
\newcommand{\emod}{Mod\xspace}
\newcommand{\ediv}{Div\xspace}
\newcommand{\elog}{Log\xspace}

\usepackage{enumitem}

\title{Efficient CTC Regularization via Coarse Labels for End-to-End Speech Translation} 

\author{Biao Zhang$^1$ \quad Barry Haddow$^1$ \quad Rico Sennrich$^{2,1}$ \bigskip\\
  $^1$School of Informatics, University of Edinburgh \\
  $^2$Department of Computational Linguistics, University of Zurich \\
  \texttt{B.Zhang@ed.ac.uk, bhaddow@ed.ac.uk, sennrich@cl.uzh.ch}
}

\begin{document}
\maketitle
\begin{abstract}

For end-to-end speech translation, regularizing the encoder with the Connectionist Temporal Classification (CTC) objective using the source transcript or target translation as labels can greatly improve quality metrics.
However, CTC demands an extra prediction layer over the vocabulary space, bringing in non-negligible model parameters and computational overheads, although this layer is typically not used for inference. In this paper, we re-examine the need for genuine vocabulary labels for CTC for regularization and explore strategies to reduce the CTC label space, targeting improved efficiency without quality degradation. We propose coarse labeling for CTC (\eplabel), which merges vocabulary labels via simple heuristic rules, such as using truncation, division or modulo (\textsc{\emod}) operations. 
Despite its simplicity, our experiments on 4 source and 8 target languages show that \eplabel with \textsc{\emod} particularly can compress the label space aggressively to 256 and even further, gaining training efficiency (1.18$\times$ $\sim$ 1.77$\times$ speedup depending on the original vocabulary size) yet still delivering comparable or better performance than the CTC baseline. 
We also show that \eplabel successfully generalizes to CTC regularization regardless of using transcript or translation for labeling.

\end{abstract}

\section{Introduction}

Developing techniques to support the translation from a source-language audio to a target-language text directly, or end-to-end (E2E) speech translation (ST), has attracted increasing attention recently due to its potential of reducing translation latency and avoiding error propagation~\cite{duong-etal-2016-attentional,berard2016listen}. However, solving this task is non-trivial because of the speech-text modality gap: one word corresponds to a stochastic sequence of speech signals that vary greatly across speakers and over contexts, which increases the learning difficulty. Recent progress on E2E ST mainly focuses on bridging this gap through the encoder-decoder framework from diverse perspectives~\cite{di2019adapting,salesky-etal-2019-exploring,zhang-etal-2020-adaptive,wang2020curriculum,han-etal-2021-learning,pmlr-v139-zheng21a}.

CTC regularization is such an approach that facilitates the modeling of translation by aligning speech representations from the encoder with discrete labels dynamically via the lens of the Connectionist Temporal Classification (CTC) objective~\cite{Graves06connectionisttemporal}. \citet{9003774} first examined the use of the source transcript as discrete labels, improving translation quality consistently across various ST settings; \citet{zhang2022revisiting} further discovered that using the target translation as labels instead can also be surprisingly effective although speech-translation pairs arguably violates CTC's monotonicity prerequisite. 
Nevertheless, these successes come at the cost of increased computational overheads and model parameters because CTC demands an extra prediction layer over its label space for probability estimation and this space is often huge -- traditionally the source or target vocabulary size~\cite{gaido-etal-2020-end}. We thus explore strategies to achieve the best of both worlds, i.e., improving the efficiency of CTC regularization without hurting its performance.

\begin{table*}[t]
    \centering
    \small
    \begin{tabular}{lcccl}
    \toprule
    \multicolumn{2}{c}{Method} & Mapping & \# Labels & \multicolumn{1}{c}{ID Sequence} \\
    \midrule
    \multicolumn{2}{c}{Genuine Labels} & \texttt{f(z) = z} & $V$ & \texttt{0,1,2,3,4,5,6,7,8} \\
    \midrule
    \multirow{4}{*}{\eplabel} & Truncation & \texttt{f(z) = min(z, L-1)} & $L$ & \texttt{0,1,2,2,2,2,2,2,2} \\
    & Modulo & \texttt{f(z) = z |\%| L} & $L$ & \texttt{0,1,2,0,1,2,0,1,2} \\
    & Division & \texttt{f(z) = |$\lfloor$|z / (V/L)|$\rfloor$|} & $L$ & \texttt{0,0,0,1,1,1,2,2,2} \\
    & Log-Scaling & \texttt{f(z) = |$\lfloor$|log(max(z,1)) * L/log(V)|$\rfloor$|} & $L$ & \texttt{0,0,0,1,1,2,2,2,2} \\
    \bottomrule
    \end{tabular}
    \caption{Overview of different mappings for \eplabel. $V$ and $L$ denote the original vocabulary size and the specified coarse label size, respectively, and $V \gg L$. $z$ is the vocabulary-space token ID starting from 0. The ID sequence is just for a toy example, and we set $V=9, L=3$ for illustration. $f(\cdot)$ shows the mapping function. \eplabel adopts different operations to reduce the label space from $V$ to $L$.}
    \label{tab:method_illustration}
\end{table*}

We address this problem by reexamining the need for genuine vocabulary labels for CTC. In contrast to CTC-based generation~\cite{Graves06connectionisttemporal}, the prediction layer in CTC regularization of ST is discarded at inference. In other words, sticking to genuine labels is computationally unnecessary. 
Since the large label space of CTC is a crucial bottleneck hindering  training efficiency, we explore ways of reducing it. We propose \textbf{Co}arse \textbf{La}beling for CTC (\eplabel) that manipulates this space by merging vocabulary labels based on simple heuristic rules.                                                 
Concretely, we map the source or target vocabulary to a pseudo label space subject to some predefined size using simple operations, such as \textit{truncation}, \textit{modulo}, \textit{division} and \textit{log-scaling} as shown in Table \ref{tab:method_illustration}.

Despite the label space being transformed, the generated coarse labels still maintain a strong correlation with their vocabulary counterparts, ensuring their informativeness for representation learning. 
We rigorously examined our method on the MuST-C~\cite{di-gangi-etal-2019-must} and the Multilingual TEDx~\cite{salesky21_interspeech} benchmarks, covering 4 source languages and 8 target languages. Across diverse settings, \eplabel successfully achieves comparable or even better translation performance than the CTC baseline but with significantly improved training efficiency (up to 1.77$\times$ speedup depending on the original vocabulary size). Our main contributions are summarized below:\footnote{Source code: \url{https://github.com/bzhangGo/zero}.}
\begin{itemize}
    \item We propose coarse labeling for CTC regularization which offers a mechanism to decouple the CTC label size from the vocabulary size; with \eplabel, a CTC-regularized model can be trained nearly as fast as a non-CTC model.
    \item We compare two types of CTC regularization for ST, i.e., using transcript or translation for labeling, and show that transcript performs better \textit{when it is available}.
    \item \eplabel delivers promising performance on 4 source and 8 target languages, and also generalizes to both types of CTC regularization.
    \item Our empirical analysis reveals that \eplabel benefits translation similarly to the CTC baseline on different aspects, including homophone translation, and seems to improve the contextualization of speech representations.
\end{itemize}

\section{Related Work}

Solving E2E ST requires techniques to mitigate the speech-text modality gap. One way is to develop advanced architectures integrating speech-specific characteristics to the encoder, such as locality modeling for the self-attention~\cite{di2019adapting,gulati20_interspeech} and adaptive speech representation grouping~\cite{salesky-etal-2019-exploring,Liu2020BridgingTM,zhang-etal-2020-adaptive}.
Another way is to leverage knowledge from other languages and/or tasks, including automatic speech recognition (ASR) and machine translation (MT) based multi-task modeling~\cite{anastasopoulos-chiang-2018-tied,dong2021listen,du2021regularizing}, cross-lingual transfer learning~\cite{inaguma2019multilingual,9004003,li-etal-2021-multilingual}, and large-scale weakly, semi- and self-supervised pretraining~\cite{Schneider2019a,ao2021speecht5,bapna2022mslam}. 
Our method contributes to E2E ST by accelerating CTC regularization with coarse labels, and is theoretically orthogonal to all the techniques aforementioned. In this study, we mainly focus on bilingual ST using triplet data alone, and leave the exploration of how our method is compatible with other setups to future work.

CTC was first proposed to handle the sequence mismatch problem between acoustic features and transcript tokens, and has been widely applied to ASR~\cite{Graves06connectionisttemporal,10.1145/1143844.1143891,7472621} and other tasks where the input sequence is longer than the output and their alignment is monotonic~\cite{10.1007/978-3-030-58517-4_11,cai2022wctc}. Recent studies also show promising results when applying CTC to non-monotonic tasks, specifically to non-autoregressive ST and MT~\cite{libovicky-helcl-2018-end,saharia-etal-2020-non,gu-kong-2021-fully,chuang-etal-2021-investigating}. 
All these methods treat the prediction layer in CTC as a generator, used to predict final outputs. By contrast,
\citet{7953075} used CTC as an auxiliary objective to improve ASR. In E2E ST, \citet{Liu2020BridgingTM}, \citet{gaido-etal-2021-ctc}, \citet{xu-etal-2021-stacked} and \citet{dong2021listen} leveraged CTC to compress speech representations to bridge the modality gap; \citet{9003774} and \citet{gaido-etal-2022-efficient} explored CTC as a regularizer using source transcript as labels, showing encouraging performance although the prediction layer is not used at inference;
\citet{zhang2022revisiting} further investigated E2E ST without transcript, and found that CTC regularization with translation as labels also works. 
Our study follows CTC regularization and extends it with \eplabel to address its training inefficiency issue. As far as we know, exploring coarse labels for CTC has never been investigated before, at least on ST. 

\section{Background: CTC Regularization}

CTC regularization improves the encoder-decoder based E2E ST by adding a CTC regularizer to the conventional translation loss~\cite{9003774,zhang2022revisiting}. Formally, given a \textit{(source speech, target translation)} pair denoted as $(X, Y)$ respectively, E2E ST with CTC regularization is optimized via the following interpolated objective:\footnote{Note we use $X, Y$ to denote the input, and their bold variants to denote the learned hidden representations.}
\begin{equation}\label{eq:ctc_reg}
    \underbrace{(1-\lambda)\mathcal{L}^{\textsc{Mle}}(Y|\mathbf{Y})}_{\text{translation loss}} + \underbrace{\lambda\mathcal{L}^{\textsc{CTC}}(Z|\mathbf{X})}_{\text{regularization}},
\end{equation}
where $\lambda$ is a hyperparameter balancing different sub-objectives. $\mathbf{X} \in \mathbb{R}^{|X|\times d}$ and $\mathbf{Y} \in \mathbb{R}^{|Y|\times d}$ denote the encoder output (or speech representation) and the decoder output, respectively. $Z$ is the label sequence for CTC, which is often either the source transcript or the target translation. $|\cdot|$ indicates the sequence length and $d$ is the model dimension. 

The translation loss $\mathcal{L}^{\textsc{Mle}}$ aims at maximizing the likelihood of observed training instances. Often, we decompose the likelihood token-wise in accordance with the autoregressive generation
\begin{align}
    \mathcal{L}^{\textsc{Mle}}(Y|\mathbf{Y}) = - \sum_{t} \log p(y_t | y_{<t}), \label{eq:mle} \\
    \text{with}\quad p(y_t | y_{<t})  \sim \text{softmax}\left(\mathbf{W}^{\textsc{Mle}}\mathbf{y}_t\right), \label{eq:softmax_layer}
\end{align}
where $y_t$ stands for the $t$-th target token. $\mathbf{y}_t \in \mathbb{R}^d$ is the $t$-th row of $\mathbf{Y}$, representing the translation prefix $y_{<t} = \{y_1, \ldots, y_{t-1}\}$. $\mathbf{W}^{\textsc{Mle}} \in \mathbb{R}^{V^\textsc{Mle}\times d}$, also called softmax embedding, is a trainable parameter, and $V^\textsc{Mle}$ is the target vocabulary size. E2E ST uses this embedding to estimate the translation probability of each target word as shown in Eq. (\ref{eq:softmax_layer})\footnote{We drop the bias term for clarity.}. At inference, the emitted probability offers direct evidence to search translation candidates.

By contrast, the regularization term $\mathcal{L}^{\textsc{CTC}}$ encourages the dynamic alignment of speech representations ($\mathbf{X}$) with their corresponding discrete labels ($Z$) through the CTC algorithm. CTC regularization also maximizes the likelihood. But different from the token-by-token formulation in Eq. (\ref{eq:mle}), CTC estimates the likelihood by marginalizing over all valid mappings between the input and output sequence~\cite{Graves06connectionisttemporal} 
\begin{align}
    \mathcal{L}^{\textsc{CTC}}(Z|\mathbf{X}) = - \log \sum_{A \in \Gamma(Z)} \prod_{k} p(a_k|\mathbf{x}_k), \label{eq:ctc} \\
    \text{with}\quad p(a_k|\mathbf{x}_k) \sim \text{softmax}\left(\mathbf{W}^{\textsc{CTC}}\mathbf{x}_k\right), \label{eq:prediction_layer}
\end{align}
where $\Gamma(Z)$ denotes the set of all valid aligned sequences.  The probability of each aligned label $a_k$ in the sequence $A$ is estimated by a prediction layer based on the corresponding speech representation $\mathbf{x}_k$ as shown in Eq. (\ref{eq:prediction_layer}). $\mathbf{W}^{\textsc{CTC}} \in \mathbb{R}^{V^{\textsc{CTC}} \times d}$ is the prediction parameter, and $V^{\textsc{CTC}} = V + 1$ denotes the CTC label size. When the source transcript (target translation) is used as labels, $V$ is the source (target) vocabulary size; the extra label is for the special \textit{blank} symbol. We refer readers to~\citet{Graves06connectionisttemporal} for more details.
Note that this predication layer will be discarded after training when CTC is purely used for regularization.

Previous studies have examined using either source transcript~\cite{9003774,gaido-etal-2022-efficient} or target translation~\cite{zhang2022revisiting} as labels for CTC regularization, but separately. In Section \ref{sec:ctc_transcript}, we will compare these two types of CTC regularization under the same setup.

\section{Coarse Labeling for CTC}

Unfortunately, CTC regularization suffers from training inefficiency. Despite CTC being efficiently addressed via dynamic programming, its prediction layer in Eq. (\ref{eq:prediction_layer}) is unavoidable. This layer introduces considerable computational overhead due to scaling linearly with the vocabulary size and also brings in large number of model parameters particularly when $V \neq V^{\textsc{Mle}}$. How to improve the efficiency, and save model parameters while retaining the performance is the focus of our study.

We draw inspiration from the fact that the prediction layer in CTC regularization is not used for inference. Thus, sticking to genuine vocabulary labels is unnecessary technically. If we could design pseudo labels in a reduced label space as alternatives to the genuine ones, that would address the inefficiency issue. Following this intuition, we propose coarse labeling for CTC (\eplabel) which formulates the pseudo label generation process as a vocabulary mapping:
\begin{equation}\label{eq:eplabel}
    f(z): \mathbb{N}_{[0\ldots V-1]} \mapsto \mathbb{N}_{[0\ldots L-1]},
\end{equation}
where $z$ denotes the original vocabulary ID. $L$ is a hyperparameter specifying the label size, and we often set $L \ll V$. This transformation decouples the label size of CTC from the vocabulary size, offering flexibility to optimize the training efficiency. One assumption behind such formulation is that the success of CTC regularization mainly comes from the inductive biases of the CTC algorithm rather than the genuineness of its labels\footnote{The inductive biases include the modeling of local structures for speech, word boundary identification and label-guided speech representation learning, etc. }, which we verified empirically through experiments.

Eq. (\ref{eq:eplabel}) merges a set of vocabulary labels into one label according to $f(\cdot)$. 
Potential mappings are many, such as grouping semantically similar words or considering phonetic similarity. But these linguistically inspired approaches often lack freedom in manipulating the label space ($L$). 
Instead, we adopt the following heuristic methods (as shown in Table \ref{tab:method_illustration}):\footnote{Note that we followed the standard practice and ranked the items in our vocabulary based on their frequency. In this study, the items are (sub)words.}
\begin{description}
    \item [Truncation (\textsc{\etru})] Learning speech representations for infrequent items is often difficult, so we merge all vocabulary labels except for the top-$(L-1)$ in \textsc{\etru}
    \begin{equation}
        f(z) = \min(z, L-1),
    \end{equation}
    with the hope that those frequent items could provide informative clues for CTC. 
    
    \item [Modulo (\textsc{\emod})] Nevertheless, infrequent items might carry crucial content information. Instead of naively collapsing them, \textsc{\emod} merges diverse labels of varying frequencies based on a fixed interval
    \begin{equation}
        f(z) = z \pmod{L}.
    \end{equation}

    \item [Division (\textsc{\ediv})] Different from items with varying frequencies, items of similar frequency often share similar linguistic properties. We explore this in \textsc{\ediv} which merges labels of similar ranks uniformly
    \begin{equation}
        f(z) = \left\lfloor{z} * \nicefrac{L}{V}\right\rfloor,
    \end{equation}
    where $\lfloor \cdot \rfloor$ denotes the floor function.
    
    \item [Log-Scaling (\textsc{\elog})] One drawback of \textsc{\ediv} is that the distribution of its coarse labels becomes badly skewed. To offset this problem, we further study a non-linear, log-scaled transformation, \textsc{\elog}
    \begin{equation}
        f(z) = \left\lfloor \log\left(\max\left(z, 1\right)\right) * \nicefrac{L}{\log(V)} \right\rfloor.
    \end{equation}
\end{description}
Note we intentionally use simple operations to keep the simplicity of \eplabel. All the above operations are trivial to implement.

Although labels generated by these operations become linguistically less meaningful, they still keep a strong correlation with their genuine vocabulary counterparts. We expect this correlation could ensure the informativeness of each coarse label and further facilitate the generalization of \eplabel to CTC regularization. We compare different operations via experiments. 

\section{Experiments} \label{sec:exp}

\paragraph{Setup} We work on two benchmarks, Multilingual TEDx~\cite{salesky21_interspeech} and MuST-C~\cite{di-gangi-etal-2019-must}, covering 4 source and 8 target languages. 
MuST-C (v1) is an English-audio based multilingual corpus, including translations from English (En) to 8 languages: German (De), Spanish (Es), French (Fr), Italian (It), Dutch (Nl), Portuguese (Pt), Romanian (Ro) and Russian (Ru). The training data for each language pair has $\sim$452 hours with about 252K utterances on average, and we use the given dev and tst-COMMON splits as the dev and test set, respectively. 
In contrast, Multilingual TEDx is a multi-source and multi-target ST corpus, containing audios in diverse languages, although its scale is relatively small. We regard this benchmark as a testbed to examine the applicability of our method to audios other than English. We report results on 6 translation directions, i.e., Es-En, Es-Pt, Fr-En, Pt-En, Fr-Es and Fr-Pt. The training data of different language pairs ranges from 25 hours (16K utterances, Fr-Pt) to 69 hours (39K utterances, Es-En), and we use the official dev and test sets for experiments. 

We focus on bilingual ST and adopt the E2E ST model following~\citet{zhang2022revisiting} which concatenates neighboring frames for downsampling followed by a variant of Transformer for translation.
We set $\lambda=0.3$ for CTC regularization. We evaluate the translation quality using (Sacre)BLEU~\cite{post-2018-call}.\footnote{Signature: \textit{BLEU+c.md+\#ref.1+s.exp+tok.13a+v.1.4.14}} We didn't perform any filtering to the test set.
All models are implemented in \textit{Tensorflow}, and trained from scratch without any ASR or MT pretraining. We refer readers to Appendix \ref{app:setup} for details on data preprocessing and training.

\subsection{Analysis on MuST-C En-De}\label{sec:ctc_transcript}

\begin{figure}[t]
  \centering
  \small
  \includegraphics[scale=0.60]{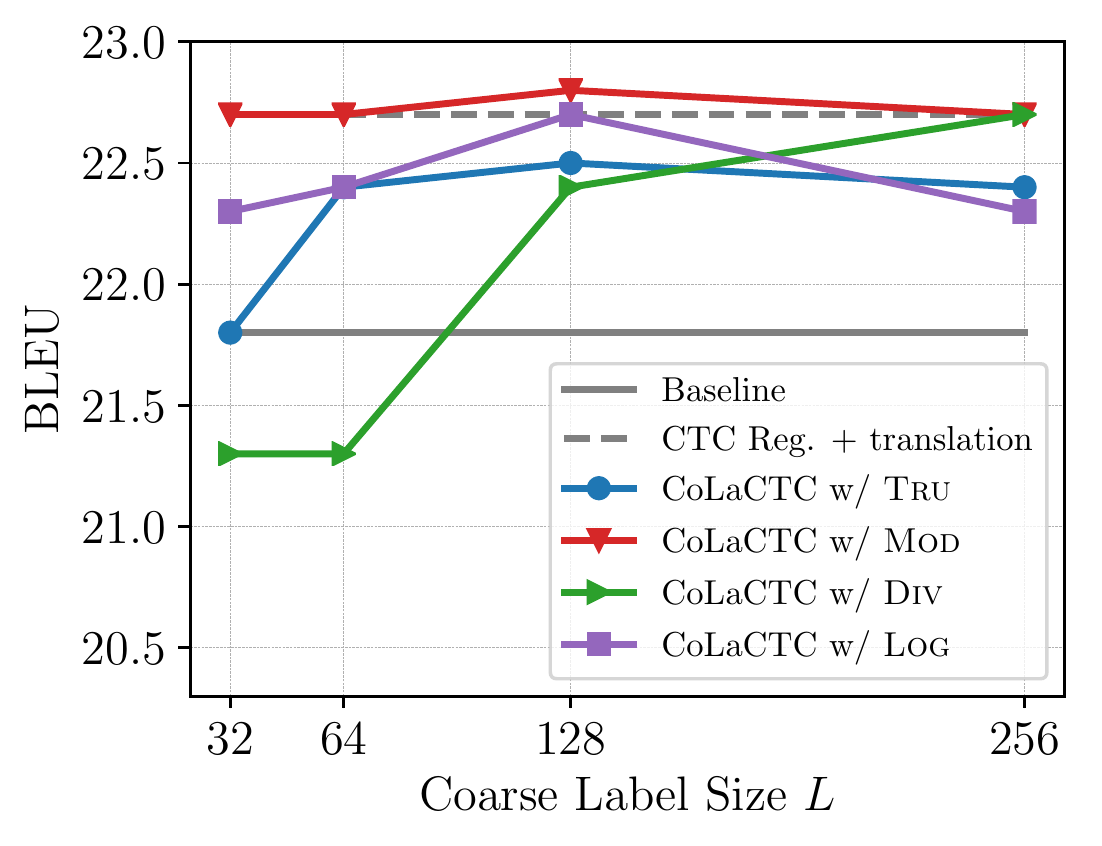}
  \caption{\label{fig:impact_L_translation} Translation results of different \eplabel methods on the MuST-C En-De test set as a function of the coarse label size $L$. \textit{Reg.}: short for regularization. Here we use \textbf{translation} labels for CTC regularization.}
\end{figure}

\paragraph{Coarse label size matters, and \textsc{\emod} performs the best.} \eplabel depends on not only the mapping function selected, but also the coarse label size specified. In general, \eplabel with a larger label size produces coarse labels closer to the genuine ones, thus behaving more robustly. We first perform ablations for \eplabel with the target translation as labels, where only speech-translation pairs are used at training~\cite{zhang2022revisiting}. 

We vary $L$ from 32 to 256, and show the results in Figure \ref{fig:impact_L_translation}. 
Different mapping functions have different properties and also show different behaviors, where the label size yields profound impacts. When $L$ is small, \textsc{\ediv} performs the worst, followed by \textsc{\elog} and \textsc{\etru} while \textsc{\emod} performs the best. With the increase of $L$, the performance difference between different mappings narrows. Label size matters, but the optimal size varies for different mappings. Under different settings, \textsc{\emod} performs the best and is most robust, nearly dominating the others. We next mainly study \textsc{\emod} for \eplabel.

\paragraph{\eplabel performs comparably to the CTC baseline.} Figure \ref{fig:impact_L_translation} also shows that \eplabel delivers comparable results to the CTC baseline when proper $L$ is applied; with \textsc{\emod} even across all tested $L$. Note both methods significantly outperform the vanilla baseline without CTC regularization.
This demonstrates that the genuineness of CTC labels matters less for CTC regularization and that our strategy -- generating coarse labels in a reduced space -- is feasible. 

\begin{figure}[t]
  \centering
  \small
  \includegraphics[scale=0.60]{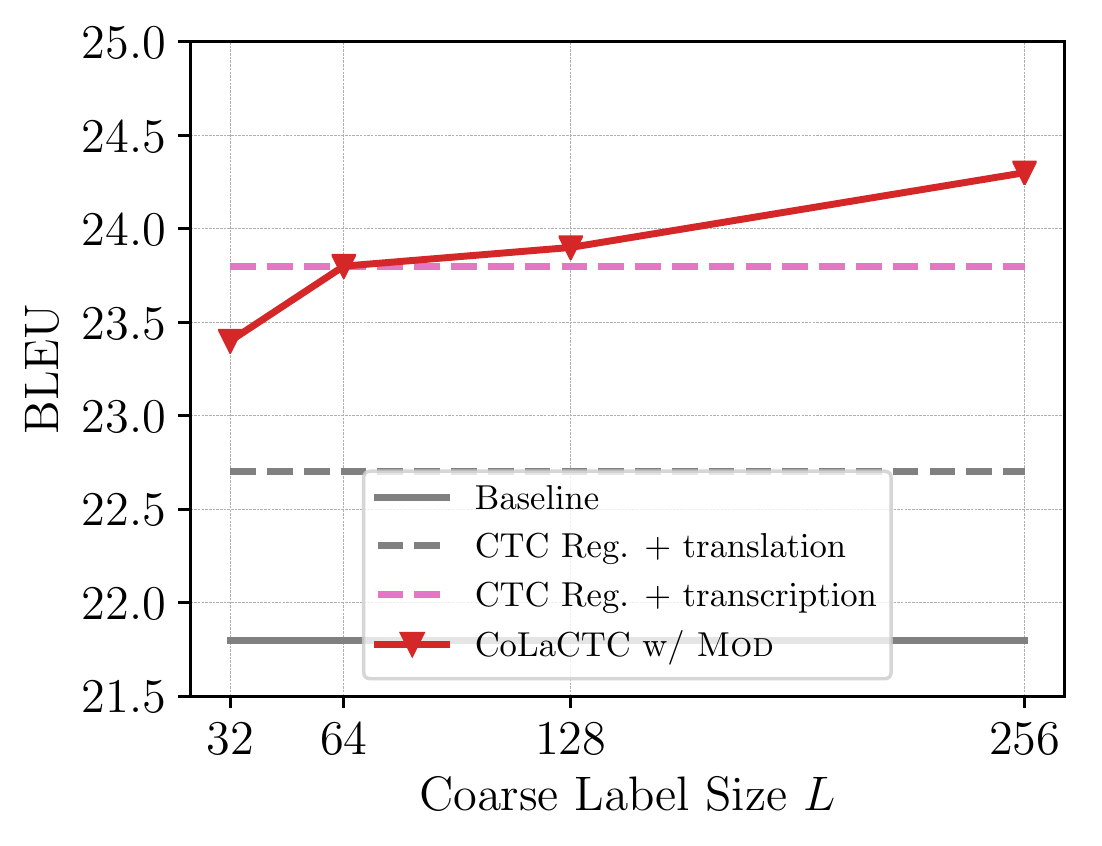}
  \caption{\label{fig:impact_L_transcription} Translation results for \eplabel with \textsc{\emod} on the MuST-C En-De test set as a function of $L$. Here we use \textbf{transcription} labels for CTC regularization.}
\end{figure}

\begin{table}[t]
    \centering
    \small
    \setlength{\tabcolsep}{5pt}
    \begin{tabular}{lccc}
      \toprule
      System & \#Param & BLEU & Speedup \\
      \midrule
      Baseline & 46.1M & 21.8 & \vphantom{25.96} 1.39$\times$ \\
      \midrule
      CTC Reg. + translation & 47.9M & 22.7 & \vphantom{36.20}1.00$\times$ \\
      ~~ + \eplabel &  46.2M & 22.7 &  \vphantom{26.06} 1.39$\times$ \\
      ~~ + share parameters & 46.1M & 22.4 & \vphantom{37.34} 0.97$\times$ \\ 
      \midrule
      CTC Reg. + transcription & 47.5M & 23.8 & \vphantom{34.21} 1.00$\times$ \\
      ~~ + \eplabel & 46.2M & 24.3 & \vphantom{26.08} 1.31$\times$ \\
    \bottomrule
    \end{tabular}
    \caption{Test results of different systems on MuST-C En-De. $L=256$ for \eplabel. \textit{\#Param}: the number of parameters. \textit{share parameters}: share parameters between $\mathbf{W}^{\textsc{Mle}}$ and $\mathbf{W}^{\textsc{CTC}}$. We perform three runs (50 steps each) to evaluate the training speedups on GeForce GTX TITAN X.}
    \label{tab:res_efficiency}
\end{table}

\paragraph{Transcript is more effective than translation as labels for CTC regularization.} Despite being effective, using translation as labels for CTC regularization violates the monotonic assumption required by CTC. CTC with transcripts is more established~\cite{9003774,gaido-etal-2022-efficient}.
We thus compare these two types of CTC regularization and explore how \eplabel generalizes.

Figure \ref{fig:impact_L_transcription} shows that using transcript instead yields substantial quality improvements ($+1.0$ BLEU), and that \eplabel with \textsc{\emod} generalizes to both settings successfully. Still, the genuineness of CTC labels matters less than their origin does! 
We observe that using 256 coarse labels works well for CTC regularization under different settings. We set $L=256$ for the following experiments.

\paragraph{\eplabel saves model parameters and greatly improves training efficiency.} CTC regularization suffers from inefficiency, which increases model parameters by about 4\% and slows the training by 39\% as shown in Table \ref{tab:res_efficiency}.
We try to solve this problem by sharing parameters between the CTC prediction layer and the softmax output layer when $V=V^{\textsc{Mle}}$ is the target vocabulary size. Unfortunately, this hurts quality and helps the training efficiency little (\textit{+ share parameters}). By contrast, \eplabel nearly recovers the efficiency sacrificed by CTC regularization, running as fast as the vanilla baseline but still retaining quality improvements. Besides, \eplabel performs similarly well with different label sequences.

We also note that the degree of inefficiency depends on the computational framework used. We re-tested different methods with PyTorch, where CTC regularization causes a 10\% decrease in training speed, much smaller than 39\%. However, the conclusion that CTC regularization leads to more trainable parameters and slower running speed, and that \eplabel overcomes this issue, still holds.

\begin{figure}[t]
  \centering
  \small
  \resizebox{\columnwidth}{!}{\includegraphics[scale=0.40]{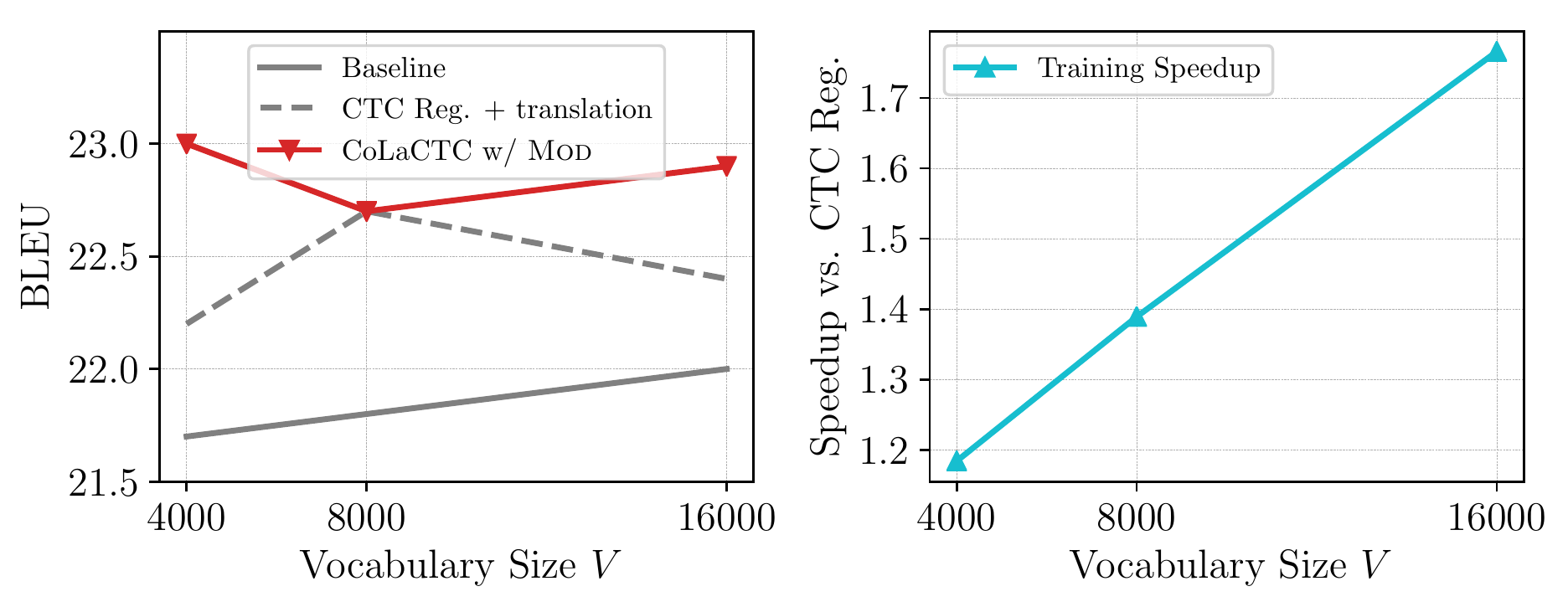}}
  \caption{\label{fig:impact_V} Test results for different methods on MuST-C En-De when varying the target vocabulary size $V$. We adopt \textbf{translation} labels for CTC regularization. $L=256$ for \eplabel.}
\end{figure}

\begin{table*}[t]
    \centering
    \small
    \begin{tabular}{lccccccccc}
      \toprule
      \multirow{1}{*}{System} & \multirow{1}{*}{De} & \multirow{1}{*}{Es} & \multirow{1}{*}{Fr} & \multirow{1}{*}{It} & \multirow{1}{*}{Nl} &\multirow{1}{*}{ Pt} & \multirow{1}{*}{Ro} & \multirow{1}{*}{Ru} & \multirow{1}{*}{Avg} \\
      \midrule
      ESPnet-ST~\cite{inaguma-etal-2020-espnet}$^\dagger$ & 22.9 & 28.0 & 32.8 & 23.8 & 27.4 & 28.0 & 21.9 & 15.8 & {25.1} \\
      Contextual Modeling~\cite{zhang-etal-2021-beyond} & 22.9 & 27.3 & 32.5 & 23.1 & 26.0 & 27.1 & 23.6 & 15.8 & 24.8 \\
      Fairseq-ST~\cite{wang-etal-2020-fairseq}$^\dagger$ & 22.7 & 27.2 & 32.9 & 22.7 & 27.3 & 28.1 & 21.9 & 15.3 & 24.8 \\
      NeurST~\cite{zhao-etal-2021-neurst} & 22.8 & 27.4 & 33.3 & 22.9 & 27.2 & 28.7 & 22.2 & 15.1 & 24.9 \\
      Wav2Vec-Transformer~\cite{han-etal-2021-learning} & 22.3 & 28.7 & 34.3 & 24.2 & 28.2 & 29.3 & 22.4 & 15.8 & 25.7 \\
      E2E-ST-JT~\cite{du2021regularizing}$^\dagger$ & 23.1 & 27.5 & 32.8 & 23.6 & 27.8 & 28.7 & 22.1 & 14.9 & {25.1} \\
      E2E-ST-TDA~\cite{du2021regularizing}$^\dagger$ & 24.3 & 28.3 & 34.6 & 24.2 & 28.7 & 30.3 & 23.4 & 15.9 & {26.2} \\
      \midrule
      Baseline & 21.8 & 27.3 & 32.3 & 22.5 & 26.6 & 27.5 & 21.8 & 14.7 & 24.3 \\
      CTC regularization + target translation & 22.7 & 28.1 & 33.4 & 23.2 & 26.9 & 28.3 & 22.6 & 15.4 & {25.1} \\
      \qquad\quad + \eplabel & 22.7 & 27.9 & 33.3 & 23.7 & 27.1 & 28.0 & 22.4 & 15.9 & 25.1 \\
      CTC regularization + source transcription & 23.8 & 28.6 & 33.9 & 24.3 & 28.3 & 29.3 & 23.3 & 16.3 & 26.0 \\
      \qquad\quad + \eplabel & 24.3 & 28.4 & 34.5 & 24.6 & 28.1 & 28.8 & 23.3 & 16.6 & \textbf{26.1} \\
    \bottomrule
    \end{tabular}
    \caption{BLEU of different systems on MuST-C \texttt{tst-COMMON}. \textit{Avg}: average score over different language pairs (translation is always out-of English). $^\dagger$: systems that might perform filtering to the test set, meaning results are not necessarily comparable. \textit{Baseline}: the model without CTC regularization; $L=256$ and \textsc{\emod} for \eplabel.}
    \label{tab:res_must_c}
\end{table*}

\begin{table*}[t]
    \centering
    \small
    \begin{tabular}{lccccccc}
      \toprule
      System & Es-En & Es-Pt & Fr-En & Pt-En & Fr-Es & Fr-Pt & Avg \\
      \midrule
      Bilingual Cascades~\cite{salesky21_interspeech} & 15.5 & 23.3 & 17.2 & 16.1 & 17.8 & 12.2 & 17.0 \\
      Bilingual E2E ST~\cite{salesky21_interspeech} & \phantom{0}7.0 & 12.2 & \phantom{0}8.9 & \phantom{0}8.1 & 10.6 & \phantom{0}7.9 & \phantom{0}9.1 \\
      Multilingual E2E ST~\cite{salesky21_interspeech} & 12.3 & 17.4 & 12.0 & 12.0 & 13.6 & 13.2 & 13.4 \\
      \midrule
      Baseline & 11.6 & 13.3 & \phantom{0}7.6 & \phantom{0}8.5 & \phantom{0}6.1 & \phantom{0}1.9 & \phantom{0}8.2 \\
      CTC regularization + target translation & 13.0 & 18.2 & 12.2 & 11.4 & 11.5 & \phantom{0}6.1 & 12.1 \\
      \qquad\quad + \eplabel & 13.3 & 19.0 & 12.1 & 12.0 & 11.2 & \phantom{0}5.0 & 12.1 \\
      CTC regularization + source transcription & 18.0 & 23.0 & 19.3 & 17.8 & 19.8 & 13.8 & \textbf{18.6} \\
      \qquad\quad + \eplabel & 17.8 & 23.1 & 19.9 & 17.5 & 19.6 & 13.3 & 18.5 \\
    \bottomrule
    \end{tabular}
    \caption{BLEU Scores on Multilingual TEDx test sets. $L=256$ and \textsc{\emod} for \eplabel.}
    \label{tab:res_mtedx}
\end{table*}
\paragraph{\eplabel performs robustly over different vocabulary sizes; larger $V$ yields higher speedups.} Apart from the coarse label size, the target vocabulary size $V$ also affects CTC regularization. Larger vocabulary shortens the target sequence but increases the CTC label space. Figure \ref{fig:impact_V} shows the impact of $V$ on \eplabel. Translation performance is highly sensitive to the vocabulary size. Using CTC regularization delivers consistent quality gains against the vanilla baseline, and \eplabel shows promising robustness, matching and even outperforming CTC regularization with genuine labels. Regarding training efficiency, the speedup of \eplabel should scale linearly with $V$ in theory when $L$ is fixed. Figure \ref{fig:impact_V} confirms this where \eplabel achieves higher speedups with larger vocabulary sizes. Particularly, the speedup reaches \textbf{1.77}{$\bm{\times}$} when $V$ is 16K, a substantial improvement.


\begin{figure}[t]
  \centering
  \small
  \includegraphics[scale=0.55]{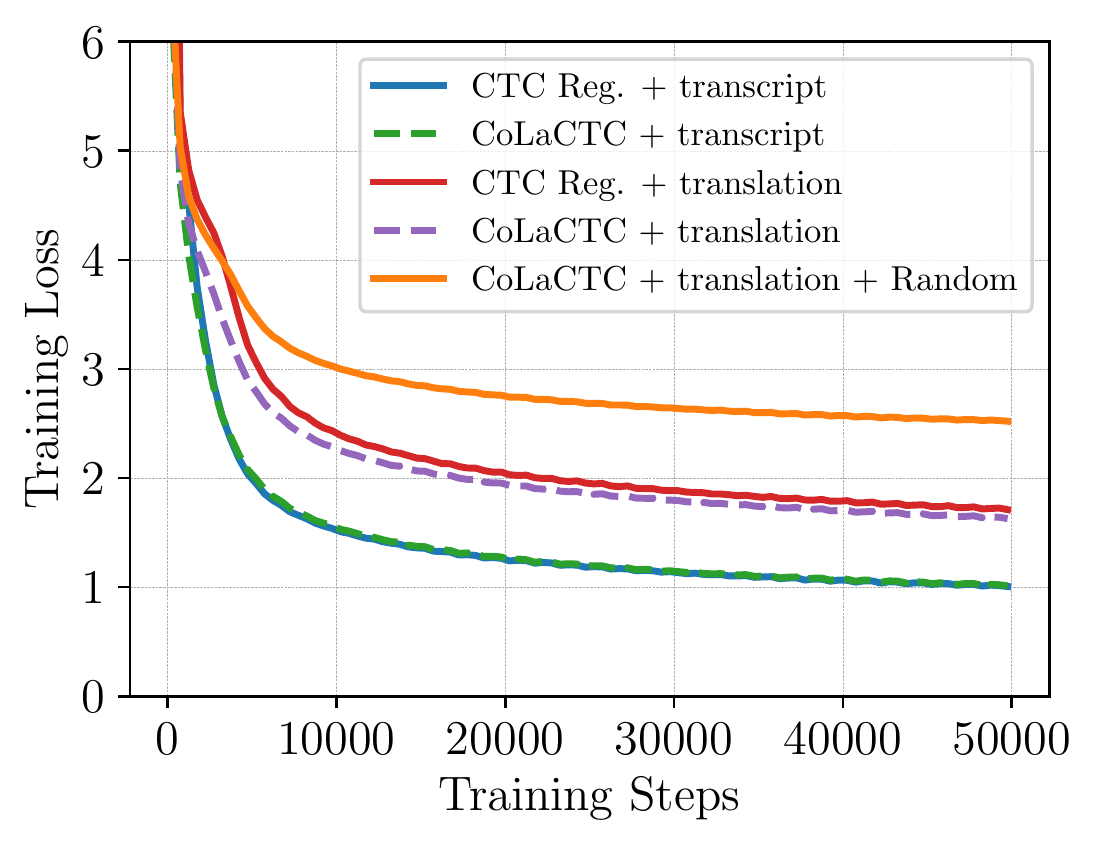}
  \caption{\label{fig:losses} Training loss as a function of training steps for different methods on MuST-C En-De. $L=256$ for \eplabel. \textit{Random} indicates using fully random labels (i.e. \texttt{f(z) = randint(0, L)}) with a label size of 256. }
\end{figure}

\paragraph{\eplabel doesn't hurt the trainability of ST models.} Would \eplabel increase the learning difficulty, which likely reduces performance? Figure \ref{fig:losses} shows that 1) \eplabel shows similar convergence to the CTC baseline using either transcript or translation for labelling; 2) the model using transcript as labels converges faster and to a better local optima than the counterpart using translation, which also explains the results in Figure \ref{fig:impact_L_transcription}; 3) random coarse labels result in inferior convergence due to their unpredictable nature. 

\paragraph{The vocabulary order of genuine labels has limited impact on \eplabel.} As shown in Eq. (\ref{eq:eplabel}), the coarse labeling in \eplabel highly relies on the order of original labels in the vocabulary. This ordering encodes word frequency, which might offer crucial clues to \eplabel and explain its success. We examine this by randomly shuffling the vocabulary, thus the original vocabulary ID is randomly changed and the ordering information is eliminated. With this shuffled vocabulary, \eplabel (\textsc{\emod}) achieves a BLEU score of 23.9 on MuST-C En-De test set, matching the performance of the CTC baseline (23.8) although underperforming the original \eplabel (24.3). \eplabel benefits from the order information but still achieves promising performance without it.

\subsection{Results on Other Languages}


\paragraph{\eplabel achieves great performance for translation out of English.} Table \ref{tab:res_must_c} summarizes the results of \eplabel on other MuST-C translation directions. The performance of \eplabel varies across different languages with both positive and negative gains. But overall, \eplabel is on par with its CTC baselines and largely outperforms the vanilla baseline without CTC regularization. On average, \eplabel delivers a BLEU score of 25.1 and 26.1 when used with translation and transcription labels, respectively, which also surpasses many strong previous studies~\cite{inaguma-etal-2020-espnet,zhao-etal-2021-neurst,zhang-etal-2021-beyond}.

\paragraph{Joint training with CTC regularization is preferable to the traditional pretraining-finetuning paradigm for E2E ST.} The current \textit{de facto} standard for training an E2E ST model is to firstly initialize it with a pretrained ASR encoder and/or MT decoder and then finetune it on ST data. Despite its effectiveness, this pipeline paradigm often consumes longer training time and inevitably complicates the optimization procedure. In contrast, joint training with CTC regularization is technically simpler and delivers comparable and even better results as shown in Table \ref{tab:res_must_c}, echoing with~\citet{gaido-etal-2022-efficient}. Note that we also re-implemented the pipeline baseline using our in-house codebase, which achieves 22.9 BLEU on MuST-C En-De, far below the joint training with transcript (23.8). 

Since \eplabel solves the inefficiency issue for CTC regularization, we would recommend using the joint training as the new standard for E2E ST, especially when only triplet training data is used.

\begin{figure*}[t]
  \centering
  \small
  \resizebox{\textwidth}{!}{\includegraphics[scale=0.40]{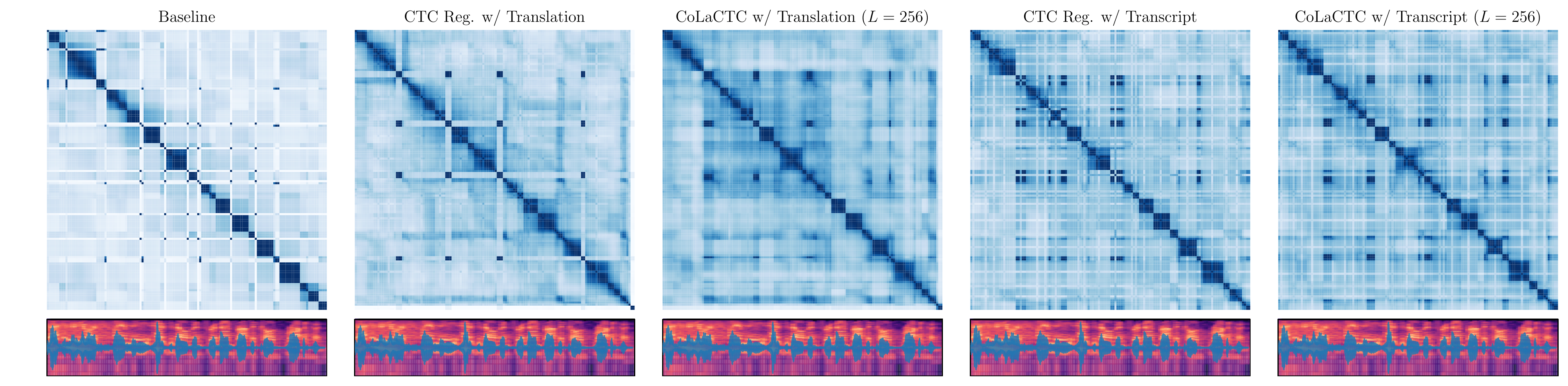}}
  \caption{\label{fig:example} Visualization of the cosine similarity of the final-layer speech encoding for different methods. \textit{Top}: cosine similarity where darker color shows higher similarity; \textit{Bottom}: speech spectrogram. The example is for the first test case in MuST-C En-De.}
\end{figure*}
\paragraph{\eplabel generalizes to ST settings other than English audios.} The above results for \eplabel are multilingual, but all use English audio from TED on the source side. To demonstrate generalization across different source languages, we conduct experiments on Multilingual TEDx and work on ST for Es, Fr and Pt. Table \ref{tab:res_mtedx} shows that \eplabel generalizes well to other source languages. In addition, CTC regularization performs much better on this benchmark, substantially outperforming the vanilla baseline by 10.4 BLEU, matching the performance of multilingual ST reported by~\citet{salesky21_interspeech}. We ascribe this success to the small scale of Multilingual TEDx where regularization techniques, like CTC Reg., often work better.

\section{Discussion}

The promising performance of \eplabel inspires us to further explore why coarse labels could work for CTC regularization. Analyzing the underlying mechanism theoretically is non-trivial. Instead, we understand this question through empirical probes, such as inspecting the change of speech representations and examining how \eplabel behaves on different translation aspects.

\paragraph{CTC regularization improves the contextualization of speech representations, so does \eplabel.} The CTC objective is directly stacked onto the encoder, then what happens to the speech representation (the final encoder output)? Figure \ref{fig:example} illustrates an example, where speech representations after applying CTC regularization (and \eplabel) become closer to each other as measured by the cosine similarity. This is further supported by the results in Table \ref{tab:res_cosine_sim}. Still, the local structure of audio, i.e. the diagonal similarity, is kept. Intuitively, the increased cosine similarity is a reflection of contextualization, and CTC regularization (and \eplabel) encourages the encoder to consider (distant) contextual clues.

\begin{table}[t]
    \centering
    \small
    \begin{tabular}{lcc}
      \toprule
      System &  Similarity \\
      \midrule
      Baseline &  0.13 \\
      CTC Reg. + translation & 0.31 \\
      ~~ + \eplabel &  0.39 \\
      CTC Reg. + transcript & 0.35 \\
      ~~ + \eplabel & 0.38 \\
    \bottomrule
    \end{tabular}
    \caption{Cosine similarity of speech representations on the MuST-C En-De test set. We report average results.}
    \label{tab:res_cosine_sim}
\end{table}

\begin{table}[t]
    \centering
    \small
    \begin{tabular}{lccccc}
      \toprule
      \multirow{2}{*}{System} & \multirow{2}{*}{MT Baseline} & \multicolumn{4}{c}{CL w/} \\ 
      \cmidrule(lr){3-6}
      & & \textsc{\ediv} & \textsc{\etru} & \textsc{\elog} & \textsc{\emod} \\
      \midrule
      BLEU & 30.5 & 10.5 & 15.1 & 22.9 & 25.6 \\
    \bottomrule
    \end{tabular}
    \caption{BLEU scores for \textit{text-to-text} translation conditioned on coarse labels on the MuST-C En-De test set. \textit{MT Baseline}: En$\rightarrow$De translation with the vanilla English input; \textit{CL w/ *}: using coarse labels instead as the source input. $L=256$. We use the standard Transformer base setting for experiments.}
    \label{tab:res_mt}
\end{table}

\begin{table}[t]
    \centering
    \small
    \setlength{\tabcolsep}{5pt}
    \begin{tabular}{lccccc}
      \toprule
      System &  Noun & Verb & Adj. & Adv. & H.Phone \\
      \midrule
      Baseline & 43.0 & 38.6 & 42.8 & 46.6 & 49.4 \\
      CTC Reg. \\
      ~~ + translation & 44.4 & 38.9 & 44.1 & 46.8 & 49.2 \\
      ~~ + \eplabel &  44.3 & 39.1 & 43.7 & 47.5 & 49.2 \\
      CTC Reg. \\
      ~~ + transcript & 45.8 & 40.6 & \textbf{46.0} & \textbf{48.3} & \textbf{51.2} \\
      ~~ + \eplabel & \textbf{46.1} & \textbf{41.1} & 45.6 & 48.0 & 51.1 \\
    \bottomrule
    \end{tabular}
    \caption{Translation accuracy of different types of source words on the MuST-C En-De test set. \textit{Adj.}, \textit{Adv.} and \textit{H.Phone} are short for adjective, adverb and homophone.}
    \label{tab:res_word_type}
\end{table}

\paragraph{Coarse labels especially produced by \textsc{\emod} preserve source semantics and are informative for translation.} The mappings considered in this study are solely based on heuristic rules. Despite improvements on ST, whether the generated coarse labels themselves are informative is still questionable. We address this concern by performing experiments on text-to-text machine translation and use the coarse label sequence as the source input. Table \ref{tab:res_mt} shows that the coarse labels encode source semantics, achieving non-trivial translation performance. In particular, using \textsc{\emod} achieves a test BLEU score of 25.6. This result still lags far behind the vanilla MT baseline (30.5), but it demonstrates the informativeness of coarse labels, and also partially explains the success of \eplabel.

\paragraph{The performance of \eplabel is robust on different types of source words.} The translation rule for different types of source words often varies greatly. Next, we examine how \eplabel generalizes to different words, including nouns, verbs, adjectives, adverbs, and homophones (specific to speech processing). We annotate the part-of-speech tag for each source sentence via Stanford POS tagger~\cite{toutanova-etal-2003-feature}, and adopt the homophone list used for contextual evaluation~\cite{zhang-etal-2021-beyond}\footnote{The list is publicly available at \url{bit.ly/3mGITEe}.}. We employ translation accuracy as the metric, approximated by the APT framework~\cite{miculicich-werlen-popescu-belis-2017-validation} where \textit{fast\_align} is used to get the word alignment~\cite{dyer-etal-2013-simple}. Table \ref{tab:res_word_type} shows the results.

\begin{figure}[t]
  \centering
  \small
  \includegraphics[scale=0.60]{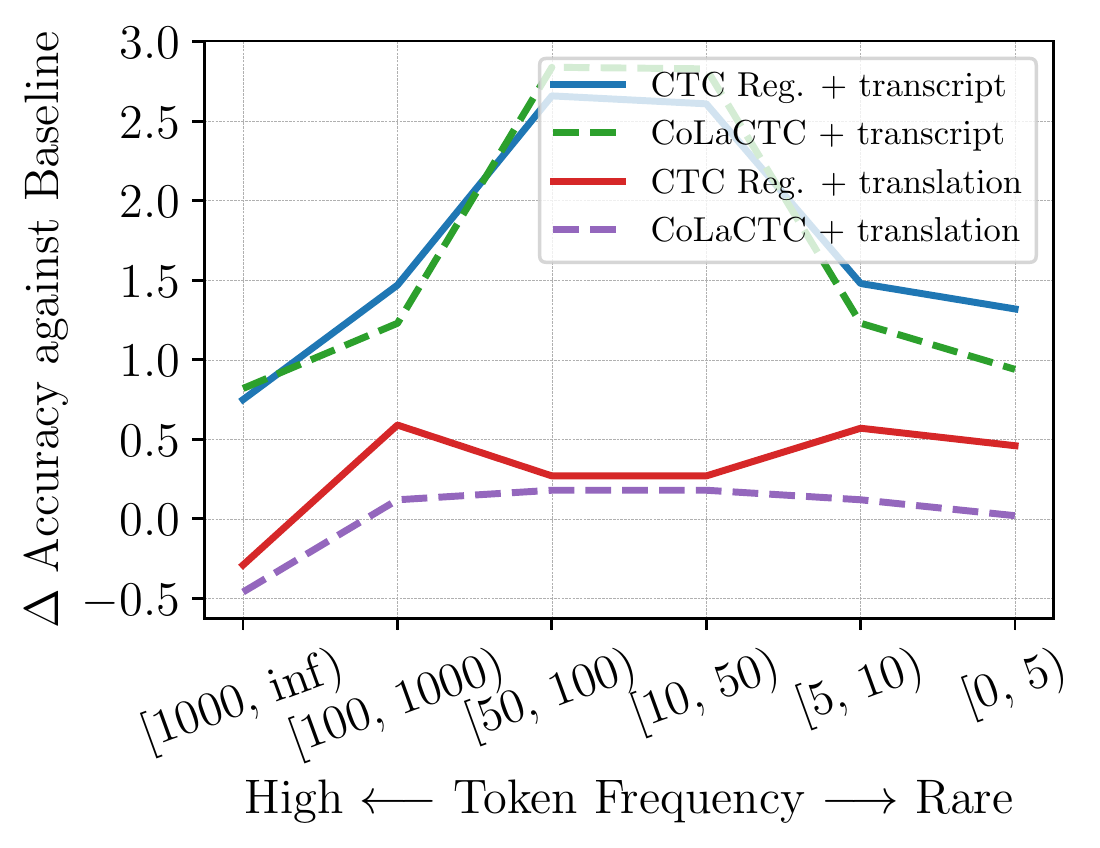}
  \caption{\label{fig:freqs} Relative gains of translation accuracy against the Baseline on MuST-C En-De as a function of token frequency ($x$-axis). We perform APT for each token group separately.}
\end{figure}

CTC regularization largely improves the translation of nouns and adjectives; using transcript as labels further benefits the translation for verbs, adverbs and homophones. Regardless of source word and CTC label types, \eplabel shows comparable (sometimes even better) performance to the CTC baseline, showing its strong generalization. Note that observations on other languages, e.g. En-Fr and En-It, are similar (see Appendix \ref{app:other_languages}). Further analysis shows that the gains by CTC regularization and \eplabel mainly come from benefiting rare-word translation, as shown in Figure \ref{fig:freqs}.

\begin{table}[t]
    \centering
    \small
    \begin{tabular}{lcccc}
      \toprule
      \multirow{2}{*}{System} &  \multicolumn{2}{c}{En-It} & \multicolumn{2}{c}{En-Fr} \\
      \cmidrule(lr){2-3}\cmidrule(lr){4-5}
        & Cov. & Acc. & Cov. & Acc. \\
      \midrule
      \citet{bentivogli-etal-2020-gender} & - & 43.3 & - & 46.0 \\
      \midrule
      Baseline & 53.3 & 65.8 & 59.3 & 64.6 \\
      CTC Reg. + translation & 56.0 & \textbf{67.2} & 59.9 & 66.5 \\
      ~~ + \eplabel & 55.9 & 67.0 & 59.6 & 65.4 \\
      CTC Reg. + transcript & 55.8 & 66.9 & 61.6 & 66.1 \\
      ~~ + \eplabel & \textbf{57.4} & 66.9 & \textbf{62.0} & \textbf{66.7} \\
    \bottomrule
    \end{tabular}
    \caption{Coverage (\textit{Cov.}) and accuracy (\textit{Acc.}) scores for gender translation on MuST-SHE En-It and En-Fr.}
    \label{tab:res_gender}
\end{table}

\paragraph{\eplabel benefits gender translation similarly to the CTC baseline.} Languages often differ in gender expression, leading to translation difficulty. We further evaluate how \eplabel handles the gender ambiguity using the MuST-SHE benchmark~\cite{bentivogli-etal-2020-gender}. Table \ref{tab:res_gender} shows that \eplabel achieves comparable performance to the CTC baseline, suggesting that using coarse labels for CTC regularization doesn't hurt its gender disambiguation ability. Besides, we observe that translation and transcript labels show similar positive effects on gender translation.

\section{Conclusion and Future Work}

In this paper, we have presented coarse labeling for CTC to address the training inefficiency issue of CTC regularization. The key idea behind \eplabel is to transform CTC labels from the vocabulary space to a specified and reduced coarse space. We adopt trivial mappings for this transformation, such as using the modulo operation. Despite its simplicity, \eplabel successfully achieves the best of both worlds -- improving the training efficiency for CTC regularization (up to 1.77$\times$ speedup) and retaining its quality benefits -- and generalizes to different types of CTC regularization. Note the training speedup scales as the vocabulary size increases. Our analysis further shows that the genuineness of CTC labels matters less than their origin.


In the future, we are interested in examining the complementarity of \eplabel with other advanced ST modeling. We will study how our method performs in a multilingual and simultaneous setup as well as ST settings with extra ASR and/or MT data.

\section*{Limitations}

While the proposed method achieves encouraging performance across diverse languages and translation setups, our understanding of why it performs so well is still limited, particularly considering the simplicity of the adopted mapping function (\textsc{\emod}). Uncovering the underlying reason behind such success might offer valuable insights to further the speech modeling, having a potential broader impact on the speech processing community.

\section*{Acknowledgements}

We thanks the reviewers for their insightful comments. This project has received funding from the European Union’s Horizon 2020 Research and Innovation Programme under Grant Agreement 825460 (ELITR), and from UK Research and Innovation (UKRI) under the UK government’s Horizon Europe funding guarantee [grant number 10039436 – UTTER]. RS acknowledges funding from the Swiss National Science Foundation (project MUTAMUR; no.\ 176727).

\bibliography{custom}
\bibliographystyle{acl_natbib}


\appendix

\begin{table*}[t]
    \centering
    \small
    \begin{tabular}{llccccc}
      \toprule
      & System &  Noun & Verb & Adj. & Adv. & H.Phone \\
      \midrule
      \multirow{5}{*}{En-It} & Baseline & 48.0 & 37.5 & 45.4 & 44.0 & 40.9 \\
      & CTC Reg. + translation & 49.6 & 38.6 & 47.2 & 45.0 & 41.7 \\
      &  ~~ + \eplabel &  49.6 & 38.7 & 48.3 & 45.4 & 42.0 \\
      &  CTC Reg. + transcript & 50.5 & \textbf{39.3} & 48.0 & 45.2 & 42.3 \\
      & ~~ + \eplabel & \textbf{51.2} & 39.0 & \textbf{49.1} & \textbf{45.7} & \textbf{43.0} \\
      \midrule
      \multirow{5}{*}{En-Fr} & Baseline & 57.2 & 51.0 & 53.4 & 53.9 & 58.1 \\
      & CTC Reg. + translation & 58.5 & 52.8 & 55.4 & 54.9 & 58.9 \\
      &  ~~ + \eplabel & 58.5 & 52.3 & 55.1 & 54.9 & 58.2 \\
      &  CTC Reg. + transcript & 59.3 & 53.0 & 54.9 & 55.1 & 59.7 \\
      & ~~ + \eplabel & \textbf{60.0} & \textbf{53.9} & \textbf{57.0} & \textbf{55.8} & \textbf{60.3} \\
    \bottomrule
    \end{tabular}
    \caption{Translation accuracy of different types of source words on the MuST-C En-Fr/En-It test set. \textit{Adj.}, \textit{Adv.} and \textit{H.Phone} are short for adjective, adverb and homophone.}
    \label{tab:res_word_type_other_langs}
\end{table*}

\section{Experimental Setting}\label{app:setup}

We preprocess texts using Moses scripts~\cite{koehn-etal-2007-moses} and adopt the byte pair encoding algorithm~\cite[BPE]{sennrich-etal-2016-neural} to handle rare tokens. In particular, we encode datasets in MuST-C and Multilingual TEDx with a BPE vocabulary size of 8K and 4K, respectively. As for audios, we adopt a sampling rate of 16KHz and filter out segments longer than 3000 frames. We extract 40-dimensional log mel-scale filterbank features for acoustic modeling with a step size of 10ms and window size of 25ms, and further augment them with their delta and delta-delta features. The final acoustic feature vector is 120-dimensional regularized by mean subtraction and variance normalization.

We focus on bilingual ST and adopt the E2E ST architecture following~\citet{zhang2022revisiting}: we use Transformer with the post layer normalization structure plus the sinusoidal positional encoding~\cite{NIPS2017_7181_attention}; we set the encoder and decoder depth to 12 and 6, respectively, and adopt the depth-scaled initialization method to stabilize the training~\cite{zhang-etal-2019-improving}; we set the model dimension to $d=256$, the feed-forward layer size to $4096$ and the number of attention head to $4$; we employ the parameterized distance penalty with $R=512$ and set $\lambda=0.3$ for CTC regularization \citep{zhang2022revisiting}. 

We train all models via Adam~\citep[$\beta_1=0.9, \beta_2=0.98$]{kingma2014adam} with a warmup step of 4K and label smoothing rate of 0.1. Samples with around 20K target subwords are scheduled into one batch for training, and we set the maximum training step for MuST-C and Multilingual TEDx to 50K and 20K, respectively. We apply dropout to residual connections and ReLU activations with a rate of 0.2. We perform checkpoint evaluation every 1K training steps on the dev set, and average the best 10 checkpoints for final testing. We use beam search for decoding, and set the beam size to $8$. We tune the length penalty for each language pair on its dev set separately.

\section{Additional Results and Analysis}\label{app:other_languages}

\paragraph{The performance of \eplabel on different types of source words generalizes to other languages.}
Table \ref{tab:res_word_type_other_langs} shows the translation accuracy of different models on En-Fr and En-It. The observation is similar to Table \ref{tab:res_word_type}, where transcript labels are more effective than translation labels for CTC regularization, and that \eplabel performs comparable to genuine labels. One exception is that CTC regularization also greatly benefits the translation of verbs and adverbs on En-Fr and En-It. These results suggest that our observation is not language-specific or caused by some random effect, but rather CTC regularization and \eplabel generalizes to different language pairs.

\end{document}